\documentclass{article}

\usepackage[preprint,eandd]{neurips_2026}

\usepackage[utf8]{inputenc}
\usepackage[T1]{fontenc}
\usepackage{hyperref}
\usepackage{url}
\usepackage{booktabs}
\usepackage{amsfonts}
\usepackage{amsmath}
\usepackage{nicefrac}
\usepackage{microtype}
\usepackage{xcolor}
\usepackage{graphicx}
\usepackage{multirow}
\usepackage{enumitem}
\usepackage{tikz}
\usetikzlibrary{arrows.meta}
\usepackage{pifont} 

\title{MemDelta: Controlled Baselines and Hidden Confounds in Agent Memory Evaluation}

\author{%
  Kuan Wang
}

\begin{document}

\maketitle

\begin{abstract}
Agent memory systems are increasingly evaluated against RAG and full-context baselines, but reported gains often mix changes in the memory method with changes in the language model, embedding model, or retrieval pipeline, making it unclear what is actually being measured. We present \textbf{MemDelta}, a controlled evaluation protocol that varies one component at a time on LongMemEval-S (500 questions, 50+ sessions, three model families). Four findings emerge: (1)~verbatim RAG matches full-context GPT-4o-mini (47.2\% vs.\ 49.8\%, $p\!=\!0.34$), but the ranking reverses across models: Gemini gains $+14$pp from full context, while Sonnet gains $+31$pp from RAG, partly because it refuses 63\% of full-context queries; (2)~swapping only the embedding model in an identical pipeline shifts accuracy by $+6.2$pp at $n\!=\!500$ ($p\!=\!0.004$), and Mem0 beats MiniLM-RAG by $+11$pp but loses to cloud-RAG by 1.2pp, so one variable flips the conclusion; (3)~agent self-memory (42\%) underperforms basic retrieval (47\%); (4)~on 2 of 6 question types ($n\!=\!88$), Mem0 matches cloud RAG (72.7\% vs.\ 73.9\%, $p\!=\!1.0$) at 50$\times$ the cost, suggesting narrow rather than general gains. We recommend memory evaluations fix embedding models across comparisons, stratify by model family, and report write-path cost before attributing gains to architecture.
\end{abstract}

\section{Introduction}

On LongMemEval-S, Mem0 appears to beat a simple verbatim-RAG baseline by a large margin when the baseline uses MiniLM embeddings: 72.7\% versus 61.4\%. This looks like an architectural win for an agent memory system over ordinary retrieval. But if we change only the embedding model (same code, same data, same retrieval logic, same prompts), the conclusion reverses: the RAG baseline reaches 73.9\%, and Mem0 loses by 1.2 percentage points. The apparent ``+11pp memory gain'' was not a property of the memory architecture. It was an embedding confound.

This kind of reversal is easy to miss because agent memory evaluations necessarily compose many moving parts. A memory system may decide what to write, how to summarize or consolidate past events, how to embed stored memories, how to retrieve them, and how to present them to a language model. Baselines such as retrieval-augmented generation (RAG) or full-context prompting are also not single points: they depend on chunking, embedding models, reranking, context formatting, model refusal behavior, and cost budgets. When two systems differ in several of these components at once, a measured improvement can be real while its cause remains unidentified.

A second example illustrates a different failure mode. For the LongMemEval-S question, ``What is the name of the playlist I created on Spotify?'' the ground-truth answer is ``Summer Vibes.'' With verbatim RAG, all three evaluated model families answer correctly. Under full-context prompting, GPT-4o-mini and Gemini also answer correctly. Sonnet, given the same full dialogue history, instead replies: ``I don't have any information about playlists you've created on Spotify.'' In this case, the retrieval method is not the only variable. The language model's behavior under long-context prompting changes the measured strength of the baseline itself.

These examples motivate the central question of this paper: when an agent memory system outperforms a baseline, what exactly improved? Recent benchmarks have made important progress by defining realistic long-horizon memory tasks, including LongMemEval~\citep{wu2024longmemeval}, MemoryAgentBench~\citep{hyun2026memoryagentbench}, MemoryArena~\citep{he2026memoryarena}, and BEAM~\citep{beam2026}. These benchmarks have helped move the field beyond short-context QA. We build on this progress. Our goal is not to argue that memory architectures cannot help, nor to replace existing benchmarks. The gap we identify is methodological: the field lacks a protocol that varies one component at a time, holds the rest fixed, and reports whether observed differences survive a significance test.

\begin{figure}[t]
\centering
\includegraphics[width=\linewidth]{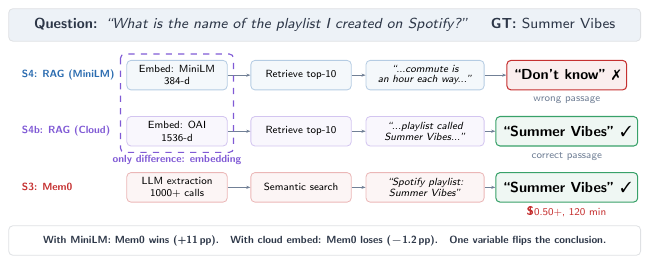}
\caption{\textbf{A measured case study: one hidden variable flips the conclusion.} The same LongMemEval-S question is processed through three pipelines. \textbf{Row~1:} MiniLM-based RAG retrieves an irrelevant passage and fails. \textbf{Row~2:} Cloud-embedding RAG (identical code, different embedder) retrieves the correct passage and succeeds. \textbf{Row~3:} Mem0's extraction pipeline succeeds but costs 1{,}000+ LLM calls. With MiniLM, Mem0 appears to win by $+11$pp. With cloud embeddings, Mem0 \emph{loses} by $-1.2$pp. Same question, same Mem0; the ``winner'' depends on which embedding the baseline used.}
\label{fig:teaser}
\end{figure}

We present \textbf{MemDelta}, a controlled evaluation protocol for agent memory systems. The protocol is deliberately simple: vary one component at a time while holding the rest of the pipeline fixed. We apply this principle to four common sources of hidden variation, shown in Figure~\ref{fig:teaser}: retrieval quality, embedding choice, language-model behavior, and write-path cost. Retrieval quality asks whether the relevant evidence is actually surfaced. Embedding choice asks whether a reported memory gain is instead a better vector representation. Model behavior asks whether the generator uses, ignores, or refuses the supplied evidence. Write-path cost asks whether a memory system's advantage comes from additional calls, summaries, or transformations that are not budget-matched against simpler baselines.

Our evaluation uses LongMemEval-S, a 500-question subset with more than 50 sessions per user and questions spanning multiple memory types. We compare verbatim RAG, full-context prompting, and an agent memory system under matched data and prompts across three model families. For RAG, we hold the retrieval code fixed and swap only the embedding model. For full-context prompting, we hold the context fixed and vary only the language model. For agent self-memory, we evaluate whether a model's own generated memories improve downstream answering over direct retrieval from the conversation history. For cost, we audit not only answer accuracy but also the extra write-time computation required to construct and maintain memories.

The resulting picture is more nuanced than a single leaderboard. In some settings, a basic RAG baseline is competitive with full-context prompting; in others, the ranking reverses across model families. A baseline that looks weak with one embedding model becomes strong with another. Agent self-memory does not automatically outperform simple retrieval. And where Mem0 is competitive with a strong RAG baseline, its advantage is concentrated in a narrow subset of question types and comes with substantially higher write-path cost. These findings do not imply that memory architectures are unnecessary. They show that uncontrolled comparisons can conflate architectural memory with retrieval engineering, embedding quality, model-specific long-context behavior, and compute budget.

\textbf{MemDelta} is therefore a measurement paper rather than a new memory architecture. We do not propose a formal causal decomposition of all components in an agent memory stack. Instead, we provide a practical protocol for making comparisons interpretable: keep the task fixed, keep the prompts fixed where possible, change one variable at a time, and report whether the conclusion survives the change. This protocol is intended to complement benchmark suites by making their results more diagnostic. A benchmark can tell us whether a system answered correctly; controlled deltas help tell us which part of the system deserves credit.

\subsection{Contributions}

\begin{enumerate}[nosep]
    \item \textbf{A controlled evaluation protocol.}
    We introduce \textbf{MemDelta}, a protocol that isolates retrieval quality, embedding choice, model behavior, and write-path cost in agent memory evaluation (\S\ref{sec:protocol}).
    \item \textbf{Controlled empirical results across model families.}
    We apply MemDelta on LongMemEval-S across three model families, showing that conclusions about RAG, full-context prompting, and agent memory change when individual components are controlled (\S\ref{sec:results}).
    \item \textbf{A cost-matched audit of Mem0 versus verbatim RAG.}
    We compare Mem0 against verbatim retrieval under matched data and prompts, auditing both accuracy and write-path cost to distinguish narrow task-specific gains from general memory improvements (\S\ref{sec:cost}).
\end{enumerate}

\section{Related Work}

\paragraph{Memory systems and benchmarks.}
Agent memory spans extraction-based (Mem0~\citep{mem0}, Cognee), verbatim (MemPalace~\citep{mempalace}), graph-based (Graphiti~\citep{graphiti}), tiered (Letta/MemGPT~\citep{packer2023memgpt}), and self-organizing (A-MEM~\citep{amem2025}) architectures, with recent work emphasizing write-path quality control~\citep{tmma2025}.
These systems are evaluated on benchmarks including LongMemEval~\citep{wu2024longmemeval}, MemoryAgentBench~\citep{hyun2026memoryagentbench}, MemoryArena~\citep{he2026memoryarena}, BEAM~\citep{beam2026}, AMA-Bench~\citep{amabench2026}, and newer interactive suites~\citep{amemgym2026, memorybench2026}.
Across all, the dominant baseline is no-memory, embedding models are rarely disclosed, and write-path cost (which can consume over 80\% of total agent execution time) is not reported.

\paragraph{Confounds in memory evaluation.}
Three bodies of work motivate our controlled protocol.
First, long-context evaluations reveal model-dependent failures: ``gold context size'' sensitivity~\citep{hiddeninhaystack2026} and semantic context modulation~\citep{semanticneedles2026} show that LLMs struggle to locate specific facts in long contexts, consistent with our Sonnet refusal findings.
Second, embedding sensitivity is well-documented in retrieval: MMTEB~\citep{mmteb2025} shows scaling laws fail for embedders, SAGE~\citep{sage2025} exposes brittleness under noise, and chunking-embedding interactions confound system comparisons~\citep{chunking2026}, yet memory benchmarks rarely control for embedding choice.
Third, the ``Benchmark Illusion''~\citep{benchmarkillusion2026} demonstrates that models with identical aggregate scores disagree on 16--66\% of items, and variable isolation studies~\citep{nebula2025} show that failing to decouple pipeline components leads to false attribution.
\citet{costperf2026} compared Mem0 against long-context LLMs but did not isolate whether the gap comes from extraction, retrieval, embeddings, or model behavior.
Cost-aware evaluation frameworks~\citep{costorchestration2026, cscr2025} increasingly argue that accuracy without cost accounting is misleading.

\paragraph{Gaps we address.}
No existing agent memory benchmark, to our knowledge, (a) isolates retrieval quality from embedding choice through controlled pairwise comparisons, (b) reports write-path cost as a first-class metric, (c) evaluates across multiple model families to test conclusion stability, or (d) provides matched-instance comparisons between extraction and confound-controlled retrieval baselines.
An extended discussion of related work is in Appendix~\ref{app:related}.

\section{Evaluation Protocol}
\label{sec:protocol}

\begin{figure}[t]
\centering
\includegraphics[width=\linewidth]{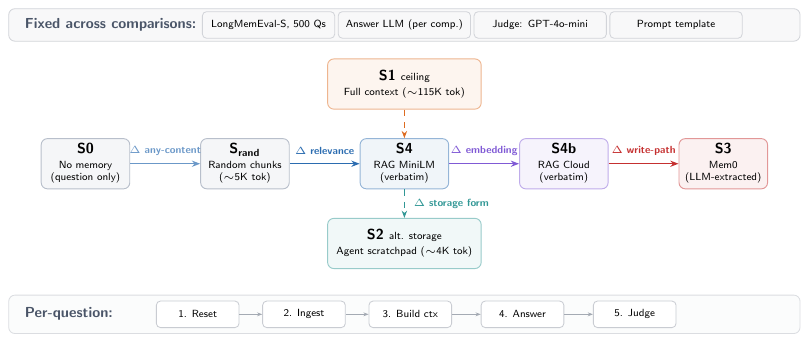}
\caption{\textbf{The MemDelta protocol.} Strategy cards connected by labeled comparison edges, each isolating one confounded variable. Top bar: components held fixed across all comparisons. Bottom: the five-step per-question procedure.}
\label{fig:protocol}
\end{figure}

Our evaluation is designed around a single principle: \textbf{vary one factor at a time}.
Rather than comparing systems that differ along multiple axes simultaneously, we construct paired conditions that isolate one variable while holding the model, judge, questions, and input history fixed (Figure~\ref{fig:protocol}).

\subsection{Strategies}

Table~\ref{tab:strategies} defines the seven strategies, ordered by write-path complexity.
Each receives the same user history and evaluation questions; they differ only in how prior information is made available to the answering model.

\begin{table}[t]
\centering
\caption{Memory strategies evaluated in MemDelta, ordered by write-path complexity.}
\label{tab:strategies}
\begin{tabular}{llp{5.8cm}}
\toprule
Strategy & Write path & Description \\
\midrule
S0 & None & Question only; no prior context. Lower bound. \\
S$_\text{rand}$ & None & Random chunks (${\sim}$5K tokens); controls for ``having text'' vs.\ ``relevant text.'' \\
S1 & None & Full conversation history (${\sim}$115K tokens) in context window. Diagnostic ceiling. \\
S4 & Embed only & Verbatim RAG: 512-tok chunks, MiniLM embeddings (384-d), top-10 retrieval. No LLM calls. \\
S4b & Embed only & Identical to S4 with OpenAI text-embedding-3-small (1536-d). Isolates embedding choice. \\
S2 & LLM (${\sim}$250 calls) & Agent scratchpad: 4{,}096-token budget, topic-organized, stale-fact overwriting. \\
S3 & LLM (1{,}000+ calls) & Mem0 v2.0 extraction pipeline. ${\sim}$120 min, \$0.50+ per instance. \\
\bottomrule
\end{tabular}
\end{table}

Three design choices deserve explanation.
First, S1 (full context) is not intended as a practical deployment strategy; it is a diagnostic control that removes retrieval as a bottleneck, allowing us to measure whether a model can use information when it is explicitly present.
Second, S4b is byte-identical to S4 except for the embedding model. The chunking, top-$k$, vector-store code, prompt template, answer model, and judge are all held fixed. Any accuracy difference between S4 and S4b is therefore attributable to embedding quality alone.
Third, S3 (Mem0) uses OpenAI embeddings internally (\texttt{text-embedding-ada-002}), while S4b uses \texttt{text-embedding-3-small}. These are different models from the same provider; the comparison approximately but not perfectly controls for embedding quality. We use S4b as the closest available controlled baseline for isolating the effect of LLM extraction on the write path, and note this imperfect match as a limitation.

\subsection{Controlled Comparisons}

\begin{table}[t]
\centering
\caption{Controlled comparisons. Each row varies one factor; all others are held fixed.}
\label{tab:protocol}
\begin{tabular}{lllr}
\toprule
Comparison & Varied factor & Held fixed & $n$ \\
\midrule
S0 $\to$ S4 $\to$ S4b & Retrieval quality & Model, judge, data & 500 \\
S$_\text{rand}$ vs S4 & Retrieval relevance & Chunking, top-$k$, model & 500 \\
S1 vs S4 $\times$ 3 models & Base-model behavior & Strategy, judge, data & 100--500 \\
S4 vs S4b & Embedding model & Pipeline code, data, model & 500 \\
S4b vs S3 (Mem0) & Write-path complexity & Embedding ($\approx$matched), model & 88 \\
S2 vs S4 & Storage control & Model, judge, data & 100 \\
\bottomrule
\end{tabular}
\end{table}

Table~\ref{tab:protocol} lists the six pairwise comparisons, each varying exactly one factor.
These are not intended as a formal causal decomposition; they are controlled probes.
If a conclusion depends on changing multiple implementation details at once, we do not attribute the effect to memory quality alone.

\subsection{Dataset and Models}

\paragraph{Dataset.}
LongMemEval-S~\citep{wu2024longmemeval}: 500 questions across 6 types (single-session-user: 70, single-session-assistant: 56, single-session-preference: 30, temporal-reasoning: 133, knowledge-update: 78, multi-session: 133).
Each instance has 39--66 sessions (mean 50.2), totaling ${\sim}$115K tokens.
The question-type diversity lets us check whether pairwise effects are uniform or concentrated in specific subsets.

\paragraph{Models.}
We run the full strategy matrix with \textbf{GPT-4o-mini} ($n\!=\!500$) and evaluate selected comparisons with \textbf{Claude Sonnet} ($n\!=\!300$--$500$) and \textbf{Gemini 2.5 Flash} ($n\!=\!100$).
The purpose of three answer models is a robustness check on the protocol itself: if a memory conclusion is robust, the relative ordering of strategies should persist across models.
If the ordering reverses when only the answer model changes, then the benchmark result is confounded by model-specific long-context behavior rather than by the memory strategy.
Sample sizes differ across models due to cost (Sonnet) and rate-limit (Gemini) constraints.
All subsets use the first $n$ instances in the benchmark's fixed ordering (by question type); we report effective $n$ per cell rather than imputing.

\subsection{Execution and Metrics}

For each (strategy, question, model) triple, we execute the same five-stage paired procedure:

\begin{enumerate}[nosep]
    \item \textbf{Reset.} Instantiate a fresh memory store; no state leaks across questions.
    \item \textbf{Ingest.} Stream the question's session history into the store using the strategy's write path. We record wall-clock time, LLM call count, and dollar cost.
    \item \textbf{Build context.} Issue the strategy's retrieval call and construct the answer prompt (S0: question only; S1: full history; S$_\text{rand}$: random chunks; S4/S4b: top-10 retrieved; S2: scratchpad; S3: Mem0 retrieval).
    \item \textbf{Answer.} Query the answer model with a fixed prompt template and temperature 0.
    \item \textbf{Judge.} Score the answer with GPT-4o-mini using a fixed binary judging prompt (correct/incorrect against ground truth).
\end{enumerate}

Because the design is paired, any judge bias constant across strategies cancels in pairwise comparisons.

\paragraph{Metrics.}
\textbf{Accuracy} with 95\% bootstrap CIs (2{,}000 resamples).
\textbf{Paired significance:} McNemar's test on per-question correctness for each comparison in Table~\ref{tab:protocol}.
\textbf{Write-path cost:} wall-clock ingest time, LLM call count, and dollar cost per instance.
Cost is not secondary: a strategy that wins by 2pp at $100\times$ the ingest cost is a different artifact than one that wins at parity, and the controlled-comparison framing is what makes that trade-off legible.

\section{Experiments}
\label{sec:results}

\subsection{Retrieval Quality: RAG Matches Full Context}

\begin{table}[t]
\centering
\caption{Main results (GPT-4o-mini, $n\!=\!500$ unless noted). McNemar vs.\ S4: * $p\!<\!.05$, ** $p\!<\!.01$, *** $p\!<\!.001$.}
\label{tab:main}
\begin{tabular}{lrccc}
\toprule
Strategy & $N$ & Accuracy & 95\% CI & vs.\ S4 ($p$) \\
\midrule
S0: No Memory         & 500 & 2.2\%  & [1.0, 3.6]   & $-45.0$pp*** \\
S\_rand: Random RAG   & 500 & 3.2\%  & [1.8, 4.8]   & $-44.0$pp*** \\
S2: Self-Memory       & 100 & 42.0\% & [32.0, 52.0]  & $-5.2$pp* \\
S4: Verbatim RAG (MiniLM)  & 500 & 47.2\% & [42.8, 51.4]  & --- \\
S1: Full Context      & 500 & 49.8\% & [45.6, 54.0]  & $+2.6$pp \\
S4b: Verbatim RAG (cloud)  & 500 & 53.4\% & [49.0, 57.8]  & $+6.2$pp** \\
\bottomrule
\end{tabular}
\end{table}

\textbf{Varied:} retrieval quality (S0 $\to$ S4 $\to$ S4b). \textbf{Fixed:} model (GPT-4o-mini), judge, data.

S0 (2.2\%) and random retrieval (3.2\%) confirm LongMemEval-S requires memory.
Verbatim RAG (47.2\%) matches full-context prompting (49.8\%), a 2.6pp gap that is \textbf{not statistically significant} ($p\!=\!0.34$).
With cloud embeddings (S4b), verbatim RAG reaches 53.4\%, significantly outperforming S4 ($p\!=\!0.004$) and numerically exceeding full context ($p\!=\!0.18$).

Agent self-memory (42\%, $n\!=\!100$) underperforms basic retrieval.
The 4{,}096-token scratchpad collapses on multi-session questions (3.3\%), where it cannot retain enough detail across 50+ sessions.
Compression is a wager: each write decides what future questions will not need.

\subsection{Model Behavior: Three Models, Three Conclusions}

\begin{table}[t]
\centering
\caption{Cross-model comparison on matched instances.}
\label{tab:cross_model}
\begin{tabular}{lrrrr}
\toprule
Model & $n$ & S4 (RAG) & S1 (Full) & $\Delta$ \\
\midrule
GPT-4o-mini        & 500 & 47.2 & 49.8 & $+2.6$pp \\
Claude Sonnet      & 300 & 44.7 & 14.0 & $-30.7$pp \\
\quad Sonnet (truncated, 47K)  & 100 & 61.0 & 38.0 & $-23.0$pp \\
Gemini 2.5 Flash   & 100 & 56.0 & \textbf{70.0} & $+14.0$pp \\
\bottomrule
\end{tabular}
\end{table}

\textbf{Varied:} base model (GPT-4o-mini, Sonnet, Gemini). \textbf{Fixed:} strategy, data, judge.

The same S1-vs-S4 comparison yields three qualitatively different conclusions.
On GPT-4o-mini, they are indistinguishable ($p\!=\!0.34$).
On Gemini~Flash, full context wins by $+14$pp.
On Sonnet, retrieval wins by $+31$pp.

A paper evaluated on a single model would draw a confident conclusion, and that conclusion would be different depending on which model was chosen.

\paragraph{Sonnet's length-driven failure.}
63\% (188/300) of Sonnet's full-context errors are explicit refusals: ``I don't see this in our conversation history,'' despite the answer being present in the 115K-token context.
A truncated-context control (${\sim}$47K tokens) reduces refusals from 63\% to 41\% and raises accuracy from 14\% to 38\%, consistent with a length-driven attention failure.
Even at 47K, Sonnet trails RAG (61\%).

The refusal pattern is not uniform across question types.
Refusal rates are highest on SS-User (70\%) and multi-session (70\%) questions, where the answer is a specific fact buried in a long history.
On SS-Preference questions, Sonnet rarely refuses (7\%) but answers \emph{incorrectly} 83\% of the time, a different failure mode where the model attempts an answer but hallucinates rather than locating the correct preference.
This taxonomy matters for benchmark design: a single ``full-context accuracy'' number hides qualitatively different failure modes that different memory strategies would address differently.

\paragraph{A concrete example.}
For the question ``What is the name of the playlist I created on Spotify?'' (ground truth: ``Summer Vibes''), all three models answer correctly with verbatim RAG, which retrieves the relevant passage.
Under full-context prompting, GPT-4o-mini and Gemini answer correctly, but Sonnet responds: ``I don't have any information about playlists you've created on Spotify.''
The answer is present in the 115K-token context; Sonnet simply cannot locate it.
With RAG, Sonnet receives only the relevant ${\sim}$5K tokens and answers correctly.
A benchmark reporting only full-context results would conclude Sonnet cannot answer this question; the failure is in the context-handling, not the memory.

\subsection{Embedding Choice: A Single Swap Rivals Architectural Gains}

\begin{table}[t]
\centering
\caption{Embedding ablation at $n\!=\!500$ (GPT-4o-mini). Identical code; only the embedding model differs.}
\label{tab:embed}
\begin{tabular}{lrrrr}
\toprule
Question Type & S4 (MiniLM) & S4b (cloud) & $\Delta$ & S1 (Full) \\
\midrule
SS-User ($n\!=\!70$)       & 75.7\% & \textbf{87.1\%} & $+11.4$ & 70.0\% \\
SS-Asst ($n\!=\!56$)       & \textbf{89.3\%} & 83.9\% & $-5.4$ & 80.4\% \\
SS-Pref ($n\!=\!30$)       & \textbf{40.0\%} & 30.0\% & $-10.0$ & 13.3\% \\
Temporal ($n\!=\!133$)     & 30.1\% & \textbf{40.6\%} & $+10.5$ & 34.6\% \\
K-Update ($n\!=\!78$)      & 62.8\% & 62.8\% & $0.0$ & \textbf{71.8\%} \\
Multi-Sess ($n\!=\!133$)   & 24.1\% & \textbf{35.3\%} & $+11.3$ & 36.8\% \\
\midrule
\textbf{Overall}           & 47.2\% & \textbf{53.4\%} & $+6.2$ & 49.8\% \\
\bottomrule
\end{tabular}
\end{table}

\textbf{Varied:} embedding model (MiniLM $\to$ cloud). \textbf{Fixed:} retrieval code, chunking, data, answer model, judge.

Switching embeddings within an otherwise identical pipeline raises accuracy from 47.2\% to 53.4\% ($+6.2$pp, $p\!=\!0.004$).
The gains are largest on temporal ($+10.5$pp), multi-session ($+11.3$pp), and single-session-user ($+11.4$pp).
The direction is not uniform: cloud embeddings underperform on SS-Asst and SS-Pref.

This is a hidden confound in memory comparisons.
A single component swap, costing nothing on the write path, produces gains comparable to those attributed to architectural changes.
Papers comparing systems with different embeddings may be confounding embedding quality with architecture.

\paragraph{Qualitative examples.}
The embedding effect is concrete: it changes which passages are retrieved, and thus which answers are correct.
For the question ``How long is your daily commute?'' (ground truth: 45 minutes each way), MiniLM retrieves a passage about commuting that mentions ``an hour each way'' from a different context, producing a wrong answer.
Cloud embeddings retrieve the passage containing the actual 45-minute detail.
For ``Where did you redeem the \$5 coupon on coffee creamer?'' (ground truth: Target), MiniLM retrieves no relevant passage and the model guesses; cloud embeddings retrieve the session mentioning Target.
In both cases, the ``architectural'' difference between a correct and incorrect system is entirely determined by which passages the embedder surfaces.

\subsection{Write-Path Cost: A Matched Comparison}
\label{sec:cost}

\begin{table}[t]
\centering
\caption{Matched comparison on 88 instances where S3 produced valid results (68 SS-User + 20 Multi-Session, GPT-4o-mini). S3 was evaluated on the first 100 dataset instances (ordered by question type in the benchmark) due to its high write-path cost (${\sim}$120 min/instance). Of 100 runs, 12 failed due to Mem0 API errors (empty responses, database lock timeouts, or extraction crashes); these are excluded rather than counted as incorrect, which may slightly favor S3.}
\label{tab:s3_matched}
\begin{tabular}{lrrc}
\toprule
Strategy & Overall & SS-User ($n\!=\!68$) & Multi-Sess ($n\!=\!20$) \\
\midrule
S1: Full Context             & 60.2\% & 72.1\% & 20.0\% \\
S4: Verbatim RAG (MiniLM)   & 61.4\% & 76.5\% & 10.0\% \\
S3: Mem0                     & 72.7\% & 88.2\% & 20.0\% \\
S4b: Verbatim RAG (cloud)   & \textbf{73.9\%} & 88.2\% & \textbf{25.0\%} \\
\bottomrule
\end{tabular}
\end{table}

\textbf{Varied:} write-path complexity (0 LLM calls $\to$ 1{,}000+). \textbf{Fixed:} embedding ($\approx$matched), answer model, judge.

On the 88 matched instances where comparison is feasible, Mem0 (72.7\%) does not outperform verbatim RAG with comparable embeddings (73.9\%, $p\!=\!1.0$, McNemar; 90\% CI on $\Delta$: [$-10.8$, $+8.4$]pp).
Note that $p\!=\!1.0$ indicates the paired disagreements are balanced, not that the systems are provably equivalent; the wide CI reflects the small sample size.
On SS-User, both score 88.2\%.
On multi-session, S3 (20\%) does not outperform S4b (25\%).

\textbf{Scope.} This comparison covers 2 of 6 question types.
On temporal reasoning and knowledge-update questions, where extraction might plausibly help most, we were unable to run S3 due to cost constraints.
We therefore do not claim extraction is generally unhelpful.
We report that, on the subset where matched comparison is feasible, extraction did not improve accuracy while incurring substantially higher write-path cost.

\begin{table}[t]
\centering
\caption{Write-path cost per instance (${\sim}$50 sessions). Cost from pilot ($n\!=\!5$); accuracy from full runs.}
\label{tab:cost}
\begin{tabular}{lrrrr}
\toprule
Strategy & Ingest time & LLM calls & Write cost & Accuracy \\
\midrule
S0: No memory    & 0\,s     & 0      & \$0      & 2.2\% \\
S4b: Verbatim RAG & ${\sim}$60\,s & 0 & \$0.01   & 53.4\% \\
S2: Self-memory  & ${\sim}$90\,min & ${\sim}$250 & \$0.34 & 42.0\% \\
S3: Mem0         & ${\sim}$120\,min & 1{,}000+ & \$0.50+ & 72.7\%$^\dagger$ \\
\bottomrule
\end{tabular}
\end{table}

{\small $^\dagger$S3 on $n\!=\!88$ (SS-User + Multi-Sess); S4b = 73.9\% on same instances.}

Table~\ref{tab:cost} reports write-path costs that accuracy-only evaluations hide.
S2 pays ${\sim}$250 LLM calls per instance yet underperforms S4b.
S3 requires 1{,}000+ calls and does not outperform S4b on matched instances.
For deployed agents ingesting thousands of interactions over time, these costs accumulate regardless of whether benchmark questions are asked.

\subsection{Per-Question-Type Analysis}

\begin{table}[t]
\centering
\caption{Per-type accuracy (\%) on GPT-4o-mini ($n\!=\!500$). S2 on $n\!=\!100$.}
\label{tab:by_type}
\begin{tabular}{lrrrrrr}
\toprule
Question Type & S0 & S\_rand & S2 & S4 & S4b & S1 \\
\midrule
SS-User ($n\!=\!70$)       & 0.0  & 5.7  & 58.6 & 75.7 & \textbf{87.1} & 70.0 \\
SS-Asst ($n\!=\!56$)       & 14.3 & 5.4  & ---  & \textbf{89.3} & 83.9 & 80.4 \\
SS-Pref ($n\!=\!30$)       & 6.7  & 3.3  & ---  & \textbf{40.0} & 30.0 & 13.3 \\
Temporal ($n\!=\!133$)     & 0.8  & 0.8  & ---  & 30.1 & \textbf{40.6} & 34.6 \\
K-Update ($n\!=\!78$)      & 0.0  & 3.8  & ---  & 62.8 & 62.8 & \textbf{71.8} \\
Multi-Sess ($n\!=\!133$)   & 0.0  & 3.0  & 3.3  & 24.1 & 35.3 & \textbf{36.8} \\
\midrule
\textbf{Overall}           & \textbf{2.2} & \textbf{3.2} & \textbf{42.0} & \textbf{47.2} & \textbf{53.4} & \textbf{49.8} \\
\bottomrule
\end{tabular}
\end{table}

The per-type breakdown reveals a structural pattern.
Retrieval excels on single-session questions (S4b: 84--87\%), where the answer typically lives in one retrievable chunk and embedding similarity reliably surfaces it.
Full-context prompting excels on knowledge-update (72\%) and multi-session (37\%) questions, where the answer requires integrating information distributed across multiple sessions that may not co-locate in any single chunk.

Cloud embeddings (S4b) close much of this gap compared to MiniLM (S4): multi-session accuracy rises from 24\% to 35\%, and temporal from 30\% to 41\%.
This suggests that some of the ``full-context advantage'' on cross-session questions is actually a retrieval-quality gap that better embeddings can address.
Full context retains a clear edge on knowledge-update (72\% vs.\ 63\%), where the model must identify which of several conflicting facts is the most recent, a reasoning task that retrieval alone does not solve.

This per-type pattern holds for Claude Sonnet (S4 single-session: 80--93\%; multi-session: 33\%), suggesting it is a structural property of the evaluation rather than a model-specific artifact.
No single strategy dominates all question types.

\subsection{Summary}

Across all four controlled comparisons, the same pattern emerges: variables that current benchmarks do not control (embedding choice, model selection, write-path cost) individually account for effect sizes as large as or larger than the architectural differences being claimed.

\begin{itemize}[nosep]
    \item Retrieval quality closes the gap from 2\% to 47--53\%, depending on the embedder.
    \item Model choice swings the S1-vs-S4 comparison by over 40pp (from Sonnet's $-31$pp to Gemini's $+14$pp).
    \item An embedding swap produces $+6.2$pp at $n\!=\!500$, enough to flip the Mem0-vs-RAG conclusion.
    \item On the matched subset, Mem0 ties cloud RAG at $50\times$ the write-path cost.
\end{itemize}

These results do not prove that memory architectures cannot improve over retrieval.
They show that the four confounds we measured are each large enough to explain or reverse the gains that current benchmarks attribute to architecture.
Until these confounds are controlled, architectural claims remain ambiguous.

\section{Conclusion}

End-to-end memory evaluations conflate retrieval quality, embedding choice, model behavior, and write-path cost.
When each is controlled separately, the picture changes.
Verbatim RAG matches full-context prompting on GPT-4o-mini, but the comparison reverses across models.
An embedding swap within an identical pipeline yields $+6.2$pp.
Agent self-memory underperforms basic retrieval.
On the matched subset where we could compare, Mem0 ties verbatim RAG at $50\times$ the cost.
These findings do not prove that memory architectures cannot improve over retrieval.
They show that current evaluations of the systems we tested cannot determine whether they do.

\paragraph{Recommended evaluation protocol.}
Based on these results, we recommend that agent memory evaluations:
\begin{enumerate}[nosep]
    \item Report against a \textbf{verbatim RAG baseline} with a named embedding model.
    \item Include \textbf{random retrieval} to separate retrieval quality from context availability.
    \item Test on \textbf{at least two model families} and report whether rank orderings are stable.
    \item Disclose \textbf{embedding models} and report embedding-swap sensitivity.
    \item Report \textbf{write-path cost} (LLM calls, time, dollars) alongside accuracy.
    \item Use \textbf{matched-instance comparisons} when evaluating costly systems on subsets.
\end{enumerate}
We release the evaluation protocol, code, and all results as open-source.
Extended discussion, including analysis of when architecture might matter, is in Appendix~\ref{app:discussion}.

\textbf{Limitations.}
Mem0 was evaluated on 88 instances covering 2 of 6 question types; temporal reasoning and knowledge-update questions, where extraction might help most, are not covered.
Sonnet full-context was evaluated on 300 of 500 instances (4 of 6 types) due to prohibitive cost.
All results are on LongMemEval-S, which uses synthetic dialogues that may favor verbatim retrieval over extraction due to higher lexical overlap; generalization to real user data is untested.
GPT-4o-mini judges all models (effect sizes are too large for judge bias to explain, but human spot-checks would strengthen confidence).
The S4b-vs-S3 comparison uses different OpenAI embedding models (text-embedding-3-small vs.\ ada-002), so the embedding control is approximate.
We report controlled contrasts, not an additive causal model; the variables may interact.


\bibliographystyle{plainnat}

\appendix
\section{Implementation Details}
\label{app:implementation}

\subsection{Retrieval Configuration}

\begin{table}[h]
\centering
\small
\caption{Retrieval hyperparameters for S4 and S4b.}
\label{tab:retrieval_config}
\begin{tabular}{ll}
\toprule
Parameter & Value \\
\midrule
Chunk size & ${\sim}$512 tokens \\
Chunk overlap & 0 (no overlap) \\
Top-$k$ retrieval & 10 \\
Similarity metric & Cosine (ChromaDB default) \\
S4 embedding & all-MiniLM-L6-v2 (384-dim) \\
S4b embedding & OpenAI text-embedding-3-small (1536-dim) \\
Vector store & ChromaDB (ephemeral, per-instance) \\
Query construction & Raw question text, no rewriting \\
Context format & Retrieved chunks concatenated, chronological order \\
\bottomrule
\end{tabular}
\end{table}

\subsection{Strategy Configurations}

\textbf{S2 (Self-Memory).}
The agent receives a system prompt instructing it to maintain a scratchpad after each session:
organize by topic (preferences, facts, events, relationships, work), overwrite stale facts, use bullet points, and keep the scratchpad under 4{,}096 tokens.
The scratchpad is passed as context at query time.

\textbf{S3 (Mem0).}
Mem0 v2.0 with default configuration.
Embedding model: OpenAI text-embedding-ada-002 (accessed via Mem0's internal pipeline).
Vector store: Qdrant (ephemeral, per-instance temp directory to avoid lock conflicts).
Top-$k$: 10.
Telemetry disabled.
Each instance requires ${\sim}$1{,}000+ LLM API calls during ingestion (one extraction call per session, plus embedding calls).

\subsection{LLM Judge}

All accuracy judgments use GPT-4o-mini with a binary prompt:
given the ground-truth answer and the model's response, output YES if the response contains the correct information, NO otherwise.
The judge does not see the question context or memory, only the ground truth and the candidate answer.
This avoids inflating scores for verbose but incorrect responses.

\subsection{Worked Examples}

\begin{table}[h]
\centering
\small
\caption{Illustrative examples showing how confounds change benchmark conclusions. Each row: same question, different outcome depending on which variable changes.}
\label{tab:examples}
\begin{tabular}{p{2.8cm}p{2.5cm}p{2.5cm}p{2.5cm}p{2.5cm}}
\toprule
Question (GT) & S4 (MiniLM) & S4b (Cloud) & GPT S1 & Sonnet S1 \\
\midrule
Daily commute? (45 min each way) & ``An hour each way'' \textcolor{red}{\ding{55}} & ``45 minutes each way'' \textcolor{green!60!black}{\ding{51}} & ``45 minutes'' \textcolor{green!60!black}{\ding{51}} & ``I don't have that info'' \textcolor{red}{\ding{55}} \\
\midrule
Where did you redeem the coupon? (Target) & ``Store not mentioned'' \textcolor{red}{\ding{55}} & ``Target'' \textcolor{green!60!black}{\ding{51}} & ``Target'' \textcolor{green!60!black}{\ding{51}} & ``I don't see info about coupons'' \textcolor{red}{\ding{55}} \\
\midrule
Spotify playlist name? (Summer Vibes) & ``Summer Vibes'' \textcolor{green!60!black}{\ding{51}} & ``Summer Vibes'' \textcolor{green!60!black}{\ding{51}} & ``Summer Vibes'' \textcolor{green!60!black}{\ding{51}} & ``I don't have info about playlists'' \textcolor{red}{\ding{55}} \\
\bottomrule
\end{tabular}
\end{table}

Table~\ref{tab:examples} illustrates how confounds change individual outcomes.
Row 1 shows the embedding confound: MiniLM retrieves the wrong passage; cloud embeddings retrieve the correct one.
Row 2 shows the same pattern.
Row 3 shows the model confound: all strategies succeed except Sonnet S1, which refuses despite the answer being in context.
In each case, the ``right'' system depends on which variable you control.

\section{Extended Discussion}
\label{app:discussion}

\subsection{What Controlled Evaluation Reveals}

The four confounds we measure are not edge cases.
Each one, individually, is large enough to explain or reverse reported architectural gains on this benchmark.
Retrieval quality accounts for the largest single effect: the gap from no-memory (2\%) to basic RAG (47\%) is $+45$pp, most of the distance to any system's absolute score.
Model choice produces the widest swing: the S1-vs-S4 delta ranges from $+14$pp (Gemini) to $-31$pp (Sonnet), a 45pp range determined entirely by which model reads the context.
Embedding choice produces gains ($+6.2$pp) comparable to those attributed to novel architectures.
And write-path cost differs by $50\times$ between Mem0 and verbatim RAG, with no measurable accuracy benefit on the matched subset.

We do not assume these factors are additive or exhaustive.
Other factors (chunk size, top-$k$, query formatting, prompt template) may also contribute.
The controlled-comparison design is not a decomposition; it is a set of probes showing that each confound alone can change the conclusion.

\subsection{Why Does Embedding Choice Matter So Much?}

The $+6.2$pp embedding effect is not uniform.
Cloud embeddings help most on temporal ($+10.5$pp) and multi-session ($+11.3$pp) questions, precisely the types that require retrieving information distributed across multiple sessions with weaker lexical overlap to the query.
Stronger semantic embeddings are better at bridging the gap between a question and a relevant passage when the surface forms differ.
Conversely, cloud embeddings \emph{hurt} on single-session-assistant ($-5.4$pp) and single-session-preference ($-10.0$pp), where MiniLM's lexical sensitivity may better match the high-overlap, single-turn nature of these questions.

This pattern has a practical implication: the ``right'' embedding model depends on the question-type distribution, which means embedding choice is not just a confound but an \emph{interaction} with task structure.
Memory benchmarks that do not disclose embedding models allow this interaction to masquerade as an architectural effect.

\subsection{When Might Architecture Matter?}

Our results do not imply that memory architectures are useless; they identify where the current evidence is ambiguous and where it is not.
Full-context prompting retains a clear advantage on knowledge-update questions (72\% vs.\ S4b's 63\%), where the model must identify which of several conflicting facts is the most recent, a reasoning task that retrieval alone does not solve.
Models with strong long-context capabilities (Gemini at 70\% overall) may genuinely benefit from seeing the complete history.
Extraction might help most on the question types we could not evaluate with Mem0 (temporal reasoning, knowledge-update), where synthesizing facts across sessions could reduce the burden on the answer model.
Whether this potential translates to measurable gains is an open question, one that our protocol is designed to answer once broader matched comparisons become feasible.
The implication is not ``stop building memory architectures'' but ``demonstrate gains under controlled conditions before claiming them.''

\section{Extended Related Work}
\label{app:related}

\paragraph{Memory systems.}
The agent memory ecosystem spans several architectural paradigms.
Extraction-based systems such as Mem0~\citep{mem0} and Cognee convert conversations into atomic facts via LLM passes during ingestion.
Verbatim systems such as MemPalace~\citep{mempalace} preserve raw text and defer interpretation to query time.
Graph-based systems such as Graphiti~\citep{graphiti} impose temporal structure on extracted facts.
Tiered systems such as Letta/MemGPT~\citep{packer2023memgpt} maintain working- and long-term memory hierarchies.
Self-organizing systems such as A-MEM~\citep{amem2025} let the agent participate in memory formation through dynamic indexing and linking.
Recent modular frameworks decompose memory into functional components (extraction, management, storage, retrieval) to enable systematic comparison, but still evaluate primarily against no-memory baselines.
The Truth-Maintained Memory Agent~\citep{tmma2025} demonstrates that write-path quality control (proactive filtering at ingestion time) is critical for long-term coherence, underscoring that write-path design deserves evaluation alongside read-path accuracy.

\paragraph{Memory benchmarks.}
LongMemEval~\citep{wu2024longmemeval} is the de facto benchmark for multi-session conversational memory (500 questions, six question types).
MemoryAgentBench~\citep{hyun2026memoryagentbench} tests cognitive competencies through incremental interactions.
MemoryArena~\citep{he2026memoryarena} measures task completion rather than retrieval, finding external memory can hurt performance.
BEAM~\citep{beam2026} evaluates at extreme scale (10M+ tokens).
AMA-Bench~\citep{amabench2026} studies memory over agent trajectories.
More recent interactive benchmarks address a limitation of static datasets: AMemGym~\citep{amemgym2026} uses LLM-simulated users to test dynamic memory updating in long-horizon conversations, and MemoryBench~\citep{memorybench2026} categorizes challenges into cognitive paradigms (object, spatial, sequential, capacity) to evaluate continual learning from feedback.
Despite this rapid progress, the dominant evaluation skeleton remains unchanged: memory systems are compared against no-memory, embedding models are rarely disclosed, and write-path cost is not reported.
Empirical profiling of memory-based workflows reveals that the memory operational cost can consume over 80\% of total agent execution time, with fact-extraction across 500 conversations requiring over 160 million tokens before a single user query is answered, a cost invisible in accuracy-only evaluations.

\paragraph{Long-context evaluation.}
\citet{costperf2026} compared Mem0 against long-context LLMs, finding a 33pp accuracy gap on LongMemEval and a cost crossover at approximately 10 turns.
However, this comparison does not isolate whether the gap comes from extraction, retrieval, embeddings, or model-specific context behavior.
Recent work provides mechanistic insight into why full-context baselines are unstable.
``Needle-in-a-haystack'' evaluations show that shorter, more specific facts are harder for LLMs to locate than longer passages, independent of overall context length, a phenomenon termed ``gold context size'' sensitivity~\citep{hiddeninhaystack2026}, and semantic needle studies show that surrounding text coherence modulates a model's ability to attend to target information~\citep{semanticneedles2026}.
These findings are consistent with our Sonnet refusal results: 63\% of full-context queries fail not because the information is absent, but because the model cannot locate it in a 115K-token haystack.

\paragraph{Embedding sensitivity in RAG.}
MTEB and BEIR show that embedding rankings shift substantially across domains, with no single model dominating all tasks.
The MMTEB extension~\citep{mmteb2025} (500+ tasks, 250 languages) reveals that scaling laws do not reliably hold for embedders: 560M-parameter instruction-tuned models frequently outperform billion-parameter variants.
The SAGE benchmark~\citep{sage2025} further exposes that top-performing embedders exhibit extreme brittleness under adversarial noise.
A comprehensive study of document chunking strategies~\citep{chunking2026} demonstrates that the interaction between chunking method and embedding model independently determines retrieval effectiveness; prior findings were ``heavily fragmented and contradictory'' precisely because this interaction was uncontrolled.

\paragraph{Evaluation methodology.}
The ``Benchmark Illusion'' phenomenon~\citep{benchmarkillusion2026} shows that models achieving identical aggregate accuracy can disagree on 16--66\% of individual items, and switching annotation models has changed treatment effect estimates by over 80\%.
Embodied AI evaluations~\citep{nebula2025} expose a ``semantic-execution gap'' where failing to isolate variables leads to false attribution of failures.
Cost-aware evaluation frameworks~\citep{costorchestration2026, cscr2025} argue that accuracy without cost accounting is misleading, with cost-spectrum routing improving the Pareto frontier by up to 25\%.

\end{document}